\documentclass{article}

\usepackage{microtype}
\usepackage{enumitem}

\usepackage{graphicx}
\usepackage{subcaption}
\usepackage{booktabs} 
\usepackage{multirow}
\usepackage{bm}
\usepackage{hyperref}

\usepackage[accepted]{icml2026}

\usepackage{amsmath}
\usepackage{amssymb}
\usepackage{mathtools}
\usepackage{amsthm}

\usepackage[capitalize,noabbrev]{cleveref}

\theoremstyle{plain}

\theoremstyle{definition}

\theoremstyle{remark}

\usepackage[textsize=tiny]{todonotes}

\icmltitlerunning{Aligning AI-driven Discovery with Human Intuition}

\begin{document}

\twocolumn[
  \icmltitle{Aligning AI-driven Discovery with Human Intuition}



  \icmlsetsymbol{equal}{*}

\begin{icmlauthorlist}
  \icmlauthor{Kevin Zhang}{mit}
  \icmlauthor{Judah Goldfeder}{columbia}
  \icmlauthor{Hod Lipson}{columbia}
\end{icmlauthorlist}

\icmlaffiliation{mit}{Massachusetts Institute of Technology, Cambridge, MA, USA}
\icmlaffiliation{columbia}{Columbia University, New York, NY, USA}

\icmlcorrespondingauthor{Kevin Zhang}{kzhang02@mit.edu}
\icmlcorrespondingauthor{Judah Goldfeder}{jag2396@columbia.edu}
\icmlcorrespondingauthor{Hod Lipson}{hod.lipson@columbia.edu}

  \icmlkeywords{Machine Learning, ICML}

  \vskip 0.3in
]



\printAffiliationsAndNotice{}  

\begin{abstract}
As data-driven modeling of physical dynamical systems becomes more prevalent, a new challenge is emerging: making these models more compatible and aligned with existing human knowledge. AI-driven scientific modeling processes typically begin with identifying hidden state variables, then deriving governing equations, followed by predicting and analyzing future behaviors. The critical initial step of identification of an appropriate set of state variables remains challenging for two reasons. First, finding a compact set of meaningfully predictive variables is mathematically difficult and under-defined. A second reason is that variables found often lack physical significance, and are therefore difficult for human scientists to interpret. We propose a new general principle for distilling representations that are naturally more aligned with human intuition, without relying on prior physical knowledge. We demonstrate our approach on a number of experimental and simulated system where the variables generated by the AI closely resemble those chosen independently by human scientists. We suggest that this principle can help make human-AI collaboration more fruitful, as well as shed light on how humans make scientific modeling choices.
\end{abstract}

\section{Introduction}\label{sec:introduction}

Recent advances in data-driven methodologies have revolutionized scientific discovery across a wide range of fields~\cite{evans2010machine,fortunato2018science}. With increasing computational power and the rise of artificial intelligence (AI), the ability to rapidly analyze vast amounts of data has significantly improved~\cite{lecun2015deep,han2018potential,akiyama2019hole}. AI algorithms now outperform humans in various tasks, including constructing predictive models~\cite{wagner2021constructions}, designing experimental protocols~\cite{coley2019robotic}, solving inverse problems~\cite{guo2024towards}, and testing causal hypotheses~\cite{king1996structure,waltz2009automating,king2004functional}. 

\begin{figure*}[t]
    \centering
    \includegraphics[width=\textwidth]{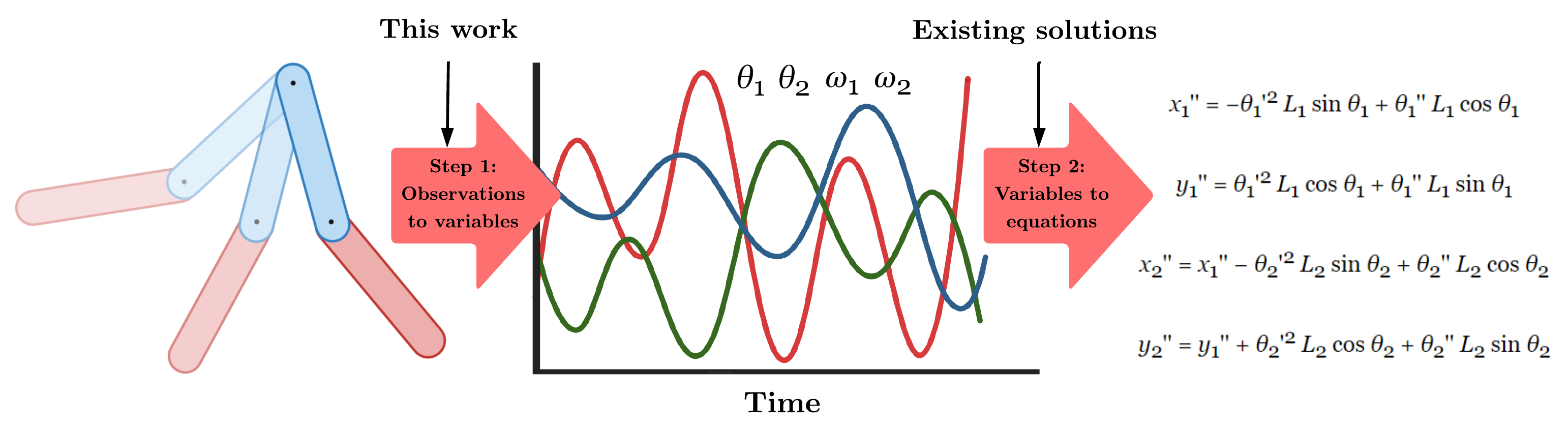}
    \caption{Overview of the automated scientific discovery process. The model first extracts relevant variables from observations of the system then derives the corresponding governing equations for the dynamics. Our work focuses on the initial state variable discovery step.}
    \label{fig:discovery}
\end{figure*}

Despite the significant progress in data-driven approaches, discovering state variables remains a challenging task. Automated scientific discovery generally involves two key steps: identifying state variables and determining the underlying dynamics governing them~\cite{kramer2023automated,cranmer2020discovering}. This process is illustrated in Figure~\ref{fig:discovery}.
By following this framework, models can extract meaningful quantities from complex systems and predict future behavior. This work focuses on the initial step of state variable discovery \textit{without prior knowledge of the system}, while simultaneously modeling the governing equations. Identifying these variables is critical for modeling new scientific systems and formulating physical laws~\cite{thompson2002nonlin,hirsch2012diffeq}. Automating this process not only deepens our understanding of complex systems that elude human intuition, but also enhances predictive modeling techniques~\cite{kramer2023automated}. Such advancements have broad implications across various fields, including decision-making~\cite{guo2021koopman,barfoot2019exactly}, biology~\cite{keil2010cardiac,jumper2021alpha}, economics~\cite{harvey1994multi}, chemistry~\cite{zhang2018protein}, 
engineering~\cite{lin2025creative},
geology~\cite{houtekamer2016kalman,carrassi2018data}, and beyond.

Significant progress has been made in modeling dynamics using methods such as sparse regression, genetic programming, and deep learning~\cite{mangan2017selection,kronberger2020identification,raissi2017diffeq,wu2019numerical,wu2019structure}, however selecting an appropriate set of state variables still remains a fundamental challenge~\cite{chen2022automated}. The difficulty stems from the non-uniqueness of state variables, even in simple systems. For example, a single pendulum can be described using either the angle and angular velocity of its arm or its kinetic and potential energies. Although numerous other combinations of variables could model a pendulum, many lack intuitive physical significance. Consequently, existing methods often rely on assumptions in the data or latent representations to identify meaningful state variables~\cite{brunton2017discovering,mrowca2018flexible,champion2019discovery,raissi2017physics,garcon2022deep,kevrekidis2002equation,chen2007modeling,bongard2007reverse,schmidt2009distilling}. In complex or novel environments where prior assumptions are absent, the problem becomes even more pressing and difficult~\cite{davies2021advancing}. This work proposes a principle to address this challenge.

Various techniques, such as dynamic mode decomposition and singular value decomposition, have been developed for discovering state variables, but these methods typically require prior knowledge of the system or underlying physical laws~\cite{kevrekidis2002equation,chen2007modeling,bongard2007reverse,schmidt2009distilling,kutz2016dynamic}. Similarly, several machine learning approaches have been proposed to infer state variables, yet they often rely on specific modeling assumptions. For example, Champion~\cite{champion2019discovery} constrained the training of an autoencoder by imposing a set of basis functions for dynamic equations, while Garcon~\cite{garcon2022deep} integrated an autoencoder with a regression module requiring partial knowledge of variables or the data-generating process. Raissi~\cite{raissi2017physics} developed a physics-informed neural network to infer dynamics, assuming certain physical laws or conditions.  While these methods are effective in scenarios with known simplifying assumptions, they become impractical in novel environments. The ability to identify state variables and derive governing equations without relying on prior modeling assumptions is crucial for understanding unfamiliar systems and advancing scientific discovery. This capability enables researchers to gain insights into unknown phenomena, particularly when theoretical predictions and empirical observations diverge.

To address the limitation of prior assumptions, Chen et al.~\cite{chen2022automated} introduced a two-stage autoencoder architecture designed to extract state variables from videos of physical systems. Their method utilizes manifold learning algorithms to estimate the intrinsic dimension (ID) of the system by employing a high-dimensional latent representation from the first autoencoder. The ID, which represents the minimum number of variables required to fully describe the system, then defines the latent dimension of a second autoencoder tasked with learning the state variables. While this approach operates without prior knowledge of the system, it struggles to produce variables with physical significance or explicit analytical expressions. This limitation stems from the non-uniqueness of the representations. Without interpretability criteria to guide the discovery process, the model tends to favor overly complex representations. Machine learning models often function as black boxes with limited transparency~\cite{lipton2017mythos,ennab2022diagnostics}, and can learn representations that do not align with human categories~\cite{lomasov2025exploring}, but for state variable discovery, interpretable representations are essential for facilitating a deeper understanding of the system~\cite{kaur2020interpreting}. 

In this work, we present a data-driven approach for learning \emph{interpretable} state variables \emph{without prior knowledge of the system}. To accomplish this, we introduce the Temporally-Informed Dynamics Encoder (TIDE), which incorporates key mathematical properties typical of physical variables into its training objective, aligning AI-generated representations with human understanding. We demonstrate that TIDE successfully extracts meaningful variables across various datasets, showing a higher correlation with human-interpretable quantities. Our main contributions are as follows: (1) we propose a method for jointly learning the dynamics, incorporating time-derivative regularization to tackle the state variable discovery problem, and (2) we learn variables that not only align more closely with human-defined quantities but also offer explicit analytical expressions and interpretable latent spaces.
\section{Problem formulation and motivation}\label{sec:problem}

\subsection*{Problem statement}

We propose a general framework for modeling dynamical systems. Following the approach in~\cite{chen2022automated}, we focus on analyzing videos of physical phenomena, though our methods can be extended to other data modalities. While images offer meaningful visual representations for humans, their high dimensionality and inherent noise make them less suitable for quantitative analysis, particularly in unknown or complex systems. Our objective is to infer the intrinsic dimension (ID) and derive a set of state variables that not only accurately capture the system's dynamics but also provide interpretative value to humans. This enables us to effectively represent, understand, and analyze the underlying dynamics.

Let $\mathcal{X}$ denote the input representation space, and consider a dataset comprising $N$ sequences $\mathcal{D} = \{\mathbf{x}_1, \ldots, \mathbf{x}_N\}$. Each sequence $\mathbf{x}_i$ consists of $M$ observations $\mathbf{x}_{i,1}, \ldots, \mathbf{x}_{i,M} \in \mathcal{X}$ sampled at a discrete time interval $\Delta t$. 
For video data, we adopt the methodology from~\cite{chen2022automated} and concatenate pairs of consecutive frames for each observation, as certain variables require information about first-order time derivatives. Thus, $\mathcal{X} \subseteq \mathbb{R}^{2 \times H \times W \times C}$, where $H$, $W$, and $C$ represent the height, width, and number of channels of each frame, respectively. We make the following assumptions about the data:
\begin{enumerate}[label=(\arabic*)]
    \item There exists a finite dimension state variable representation space $\mathcal{S}$ and a bijective function $f: \mathcal{S} \to \mathcal{X}$ such that for all $\mathbf{x} \in \mathcal{X}$, there exist unique state variables $\mathbf{z} \in \mathcal{S}$ that generate the input $\mathbf{x} = f(\mathbf{z})$.

    \item There exists a deterministic time evolution operator $g: \mathcal{S} \to \mathcal{S}$ such that $\mathbf{z}_{i,j+1} = g(\mathbf{z}_{i,j})$.

    \item $\Delta t$ is sufficiently small such that consecutive observations over time do not have large changes in the state variable representation.
\end{enumerate}
Assumptions 1 and 2 assert that there exists a finite set of state variables that evolves deterministically, while assumption 3 states that state variable representation is stable. This assumption is generally weak and is typically satisfied by efficient data measurement or high video frame rate. 

Our goal is to fully characterize the system and predict its dynamics. To achieve this, it is sufficient to determine the intrinsic dimension (ID) and then specify $f$ and $g$. Note that assumptions 1 and 2 imply the existence of an evolution operator in the input space, $g': \mathcal{X} \to \mathcal{X}$, defined as $g' = f \circ g \circ f^{-1}$. However, modeling dynamics directly in terms of pixels is both overly complex and uninformative, so we choose to model $f$ and $g$ separately. We employ a variational autoencoder (VAE) to learn $f$, while simultaneously optimizing a feed-forward network (referred to as the dynamics module) to learn $g$. This approach avoids assuming prior knowledge of the system or imposing functional constraints, making it suitable even for unknown environments. Furthermore, we do not enforce statistical dependencies between state variables or restrict the governing equations to parametric forms via basis functions, as such assumptions are not universally applicable. By avoiding these limitations, our model offers a more general framework compared to previous methods.

\subsection*{Aligning state variables}

While identifying state variables is crucial, it is equally important to prioritize those that offer interpretive value. However, assessing the alignment between human-understandable and model-learned latent variables without prior knowledge of the system poses a challenge. To address this, consistent with the principle that strong domain-specific priors are necessary for specialized scientific tasks\cite{goldfeder2026ai}, we propose three desirable qualities for typical human-interpretable state variables: they should accurately represent the system, have simple governing equations, and be infinitely differentiable over time almost everywhere. We incorporate these properties by training an autoencoder to learn the representation, jointly optimizing a dynamics module, and applying time-derivative regularization, respectively.

\begin{figure*}[t!]
    \centering
    \centering
    \includegraphics[width=0.73\linewidth]{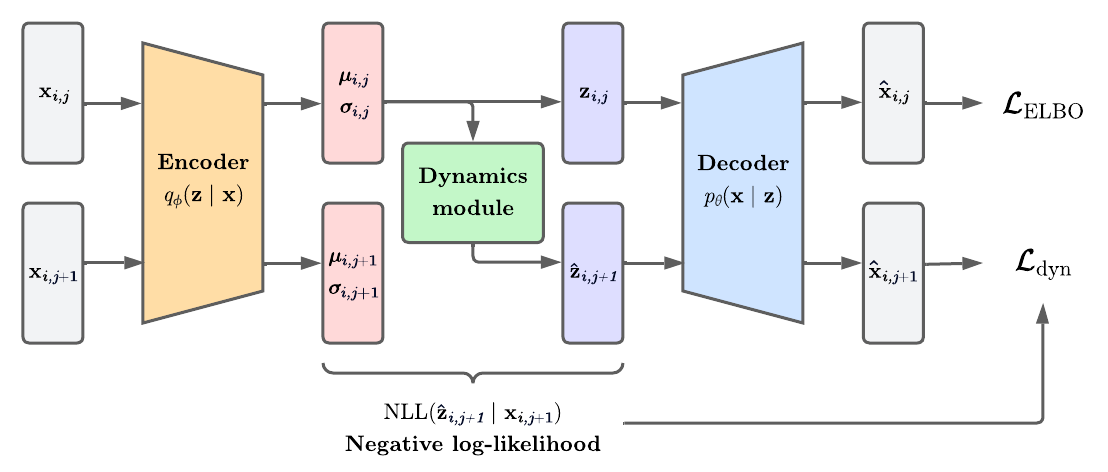}
    \caption{Schematic of the TIDE architecture for interpretable state variable discovery. The VAE backbone consists of an encoder $q_\phi$ which maps the input $\mathbf{x}_{i,j}$ to $\mathcal{N}(\bm{\mu}_{i,j}, \bm{\sigma}^2_{i,j})$  and a decoder $p_\theta$ that reconstructs the input from $\mathbf{z}_{i,j} \sim q\phi(\mathbf{z} \mid \mathbf{x}{i,j})$. A dynamics module, $h_\text{dyn}$, is integrated to predict the next latent representation $\hat{\mathbf{z}}_{i,j+1} = h_\text{dyn}(\bm{\mu}_{i,j})$ by maximizing the likelihood. The overall training objective is a weighted combination of the ELBO $\mathcal{L}_\text{ELBO}$, dynamics loss $\mathcal{L}_\text{dyn}$, and time-derivative regularization $\mathcal{L}_\text{reg}$.}
    \label{fig:architecture}
\end{figure*}
\section{Methods} \label{sec:methods}


\subsection*{Model architecture}

We present the components of the Temporally-Informed Dynamics Encoder (TIDE), which includes a variational autoencoder (VAE) backbone, dynamics module, and time-derivative penalty. Our approach uses TIDE networks to compress high-dimensional pixel data into a low-dimensional set of state variables, following a two-step process similar to~\cite{chen2022automated}.

We use a variational autoencoder (VAE) to encode the input data into a latent representation to model the function $g$. VAEs are a class of generative models comprising an encoder and decoder parameterized by $\theta$. The encoder compresses the input $\mathbf{x}$ into a low-dimensional latent representation $\mathbf{z}$, while the decoder reconstructs the data by optimizing a maximum likelihood objective. Due to the intractability of the posterior $p_\theta (\mathbf{z})$, VAEs approximate it by variational inference. Using a Gaussian variational family, we learn an encoder $h_\phi$, parameterized by $\phi$, to approximate the posterior 
\[ q_\phi (\mathbf{z} \mid \mathbf{x}) = \mathcal{N}(h_\phi^{\bm{\mu}}, \text{diag}(h_\phi^{\bm{\sigma}^2})) \approx p_\theta (\mathbf{z} \mid \mathbf{x}) \] 
by maximizing the evidence lower bound (ELBO):
\[ \mathcal{L}_\text{ELBO}(\beta) = \mathbb{E}_{\mathbf{x}} [\mathbb{E}_{q_\phi} [\log p_\theta(\mathbf{x} \mid \mathbf{z})] - \beta D_{KL}(q_\phi (\mathbf{z} \mid \mathbf{x}) \parallel p_\theta (\mathbf{z}))], \]
where $\beta$ controls the latent capacity as introduced in~\cite{higgins2017betavae}.

We jointly optimize a dynamics module $h_\text{dyn}$ integrated within the VAE to model the time evolution operator $g$. This module is trained to maximize the likelihood of the next observation. For input $\mathbf{x}_{i,j}$, let $\bm{\mu}_{i,j}$ and $\bm{\sigma}_{i,j}$ denote the mean and standard deviation of $q_\phi (\mathbf{z}_{i,j} \mid \mathbf{x}_{i,j})$, respectively. Let $\hat{\mathbf{z}}_{i,j+1} = h_\text{dyn}(\mathbf{x}_{i,j})$ represent the predicted latent state at the next time step. The dynamics objective is to maximize:
\begin{gather*}
    \mathcal{L}_\text{dyn} (\lambda_1) = \mathbb{E}_{\mathbf{x}} \left[ \log p_\theta(\mathbf{x}_{i,j+1} \mid \hat{\mathbf{z}}_{i,j}) + \lambda_1 \log q_\phi(\hat{\mathbf{z}}_{i,j} \mid \mathbf{x}_{i,j+1}) \right].
\end{gather*}
We introduce a hyperparameter $\lambda_1$ to balance the relative weight between maximizing the data likelihood and latent likelihood. The dynamics module is implemented as a fixed-width, three-layer feed-forward neural network (FFN) to maintain simplicity and limit the complexity of the governing dynamics. Alternatively, the dynamics module could be trained using symbolic or sparse regression methods (e.g., SINDy~\cite{brunton2017discovering}). However, such approaches inherently constrain the dynamics by relying on predefined basis or candidate functions. By using a neural network, we achieve greater generality. The architecture is illustrated in Figure~\ref{fig:architecture}.

\subsection*{Time-derivative regularization}

We introduce time-derivative regularization to encourage smooth state variables by applying a penalty term based on the magnitudes of discrete derivatives. To prevent global contraction of the latent space, we normalize the latent vectors along each dimension independently using min-max normalization. Given an input $\mathbf{x}$, let $\bm{\overline{\mu}}$ represent the normalized mean. The discrete $n$-th derivative at $\bm{\overline{\mu}}_{i,j}$ is defined as $\bm{\overline{\mu}}_{i,j}^{(n)} = \bm{\overline{\mu}}_{i,j+1}^{(n-1)} - \bm{\overline{\mu}}_{i,j}^{(n-1)}$, where $\bm{\overline{\mu}}_{i,j}^{(0)} = \bm{\overline{\mu}}_{i,j}$. The time-derivative regularization $\mathcal{L}_\text{reg}$ aggregates the first $n$ discrete derivatives. Since higher-order derivatives tend to vanish due to the lack of renormalization, each term is geometrically scaled by a factor $\omega$. The regularization penalty is expressed as follows:
\[ \mathcal{L}_\text{reg} = \mathbb{E}_{\mathbf{x}_i \in \mathcal{D}} \left[ \sum_{d=1}^n \omega^d \,  \mathbf{\overline{\bm{\mu}}}_i^{(d)} \right] \]
We use $n = 4$ and $\omega = 5$ in all experiments, as these values yield meaningful state variables. The final objective is a weighted combination of the ELBO, dynamics loss, and the regularization term.
\[ \mathcal{L}_\text{TIDE} = -\mathcal{L}_\text{ELBO} (\beta) - \mathcal{L}_\text{dyn} (\lambda_1) + \lambda_2 \mathcal{L}_\text{reg}. \]
We introduce a hyperparameter $\lambda_2$ to control the weight of the regularization term and refer to our model architecture, comprising the VAE backbone, dynamics module, and time-derivative regularization, as a TIDE network.

\subsection*{Intrinsic dimension estimation}

Accurate estimation of the intrinsic dimension (ID) is essential for efficient and precise modeling. In~\cite{chen2022automated}, the authors observed that directly reducing the bottleneck dimension of an autoencoder does not reliably recover the ID, as reconstructions begin to deteriorate before reaching the true value. Instead, they use a manifold learning algorithm to estimate the ID. Building on their methodology, we train a TIDE network to generate high-dimensional latent representations $\mathbf{y}_{i,j} \in \mathbb{R}^{64}$ and then apply the Dimensionality from Angle and Norm Concentration (DANCo) algorithm~\cite{ceruti2014danco} to the latent space. This provides a maximum likelihood estimate of the ID based on the distance and angle between pairs of normalized $k$-nearest neighbors. The estimated ID serves as the latent dimension for a second TIDE network, which is used to predict the final state variables. The second network is trained on top of the first (which is kept frozen) to learn state variables $\mathbf{z}_{i,j} \in \mathbb{R}^\text{ID}$ by reconstructing both the data $\mathbf{x}_{i,j}$ and the intermediate latent representation $\mathbf{y}_{i,j}$, where the additional likelihood term for $\mathbf{y}_{i,j}$ is weighted by a hyperparameter $\lambda_3$.

\begin{figure*}[t!]
    \centering
    \includegraphics[width=0.9\textwidth]{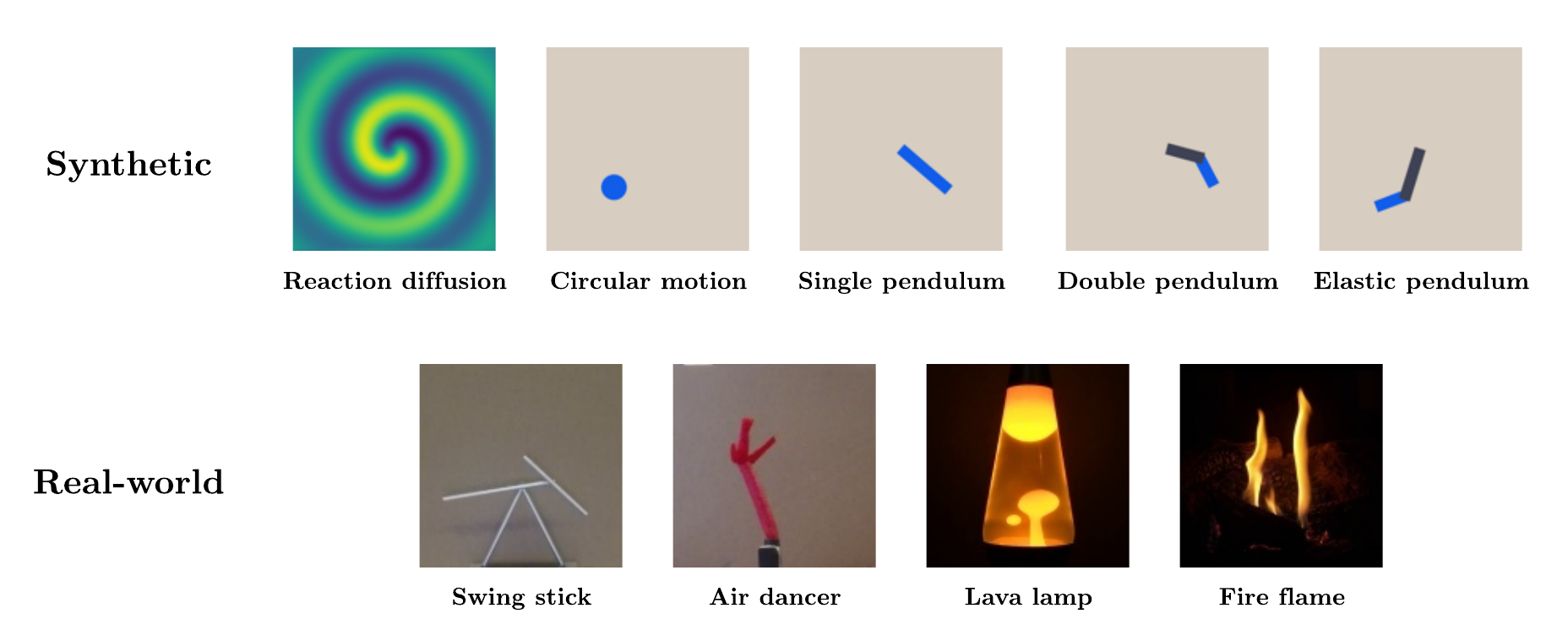}
    \caption{Grid showing the nine evaluation video datasets.}
    \label{fig:datasets}
\end{figure*}

\subsection*{Symbolic regression}

For systems with known standard human state variables, we derive an expression for the predicted state variable representation using PySR~\cite{cranmer2023pysr}, a symbolic regression library built on multi-population genetic algorithms. We restrict the search space to sine, addition, subtraction, and multiplication operations. These functions prove sufficient for accurate analytical reconstruction, as adding other elementary functions does not significantly improve the final fit.

\subsection*{Dataset}

We evaluate our model on nine benchmark datasets from~\cite{chen2022automated}, consisting of five simulated systems (reaction-diffusion, circular motion, single pendulum, double pendulum, elastic pendulum) and four real-world systems (swing stick, air dancer, lava lamp, fire flame). Figure~\ref{fig:datasets} illustrates these nine evaluation settings.
\section{Results}\label{sec:results}
To assess our model, we establish specific evaluation metrics then compare our approach and the methodology in~\cite{chen2022automated}. Our model accurately learns the dynamics and an interpretable latent representation having higher correlation with human variables. Additionally, we derive analytical expressions for the learned state variables, demonstrating they are meaningful and recoverable.

\subsection*{Evaluation metrics}
To assess variable interpretability, we use two metrics: mutual information (MI) and analytical mean squared error (AMSE). MI serves as an information-theoretic measure of correlation between model and human state variables. For two random variables $X$ and $Y$, the MI quantifies the information gain about one variable by observing the other:
\[ \text{MI}(X,Y) = \iint P(x,y) \log \left( \frac{P(x,y)}{P(x) P(y)} \right) \,dx \,dy. \]
We estimate the probability densities using Gaussian kernel density estimation.

In addition to MI, we calculate the mean squared error between the model’s state variables and their analytical fit. State variables that yield accurate and simple analytical expressions are typically more interpretable and aligned with human understanding. This metric, referred to as AMSE, quantifies interpretability by measuring the accuracy of the symbolic expression.

\begin{table*}[t!]
    \caption{Intrinsic dimension (ID) estimates for the nine benchmark datasets from both our model (TIDE) and the baseline (NSV), including the ground truth value where available. The mean and standard deviation of the ID are reported across three random training splits. TIDE consistently provides more precise estimates, which we attribute to the incorporation of time-derivative regularization.}
    \renewcommand{\arraystretch}{1.5}
    \centering
    \begin{tabular}{cccc}
    \hline & \\[-3.0ex]
    \textbf{Dataset} & \multicolumn{1}{c}{\textbf{TIDE ID (ours)}} & \multicolumn{1}{c}{\textbf{NSV ID (baseline)}} & \multicolumn{1}{c}{\textbf{Ground truth ID}} \\ & \\[-3.0ex] \hline & \\[-3.0ex]
    Reaction diffusion & $2.17 \pm 0.07$ & $2.17 \pm 0.16$ & 2 \\
    Circular motion & $2.13 \pm 0.02$ & $2.10 \pm 0.03$ & $2$ \\
    Single pendulum & $2.17 \pm 0.01$ & $2.15 \pm 0.03$ & $2$ \\
    Double pendulum & $4.02 \pm 0.03$ & $3.52 \pm 0.08$ & $4$ \\
    Elastic pendulum & $6.00 \pm 0.04$ & $4.46 \pm 0.04$ & $6$ \\
    Swing stick & $4.26 \pm 0.59$ & $3.86 \pm 0.09$ & $4$ \\
    Air dancer & $3.60 \pm 0.33$ & $4.29 \pm 0.24$ & Unknown \\
    Lava lamp & $4.99 \pm 0.35$ & $5.05 \pm 0.32$ & Unknown \\
    Fire flame & $8.01 \pm 0.01$ & $10.39 \pm 0.75$ & Unknown \\ & \\[-3.0ex] \hline
    \end{tabular}
    \label{tab:id-estimates}
\end{table*}

\subsection*{Intrinsic dimension estimation}
Accurately estimating the intrinsic dimension is a crucial step in the state variable discovery process, as it determines the latent dimension for the network that will subsequently learn the state variables. In Table~\ref{tab:id-estimates}, we compare the ID estimates from our model (TIDE) with the baseline Neural State Variables (NSV) approach from~\cite{chen2022automated}. Ground truth ID values, where available, are also included. For the baseline, we follow their outlined methodology to calculate the estimate. To ensure a fair comparison, we use $\mathbb{R}^{64}$ as the intermediate latent space for all ID estimation algorithms. Each method is trained on three random training, validation, and test splits, and we report the mean and standard deviation.

For datasets where the ID is known, our model accurately identifies the necessary number of state variables. TIDE estimates tend to align more closely with the ground truth, likely due to the time-derivative regularization, which encourages a smooth and continuous latent representation, resulting in more precise estimates. Further exploration of the latent space is provided in the subsequent analysis. For datasets with unknown ID, both models generally produce consistent estimates, although TIDE predicts a slightly smaller value for the fire flame dataset. To determine the latent dimension for the subsequent model, we use the ground truth value when available, otherwise rounding the predicted ID to the nearest integer.

\begin{figure*}[t]
    \centering
     \begin{subfigure}[t]{0.48\textwidth}
         \centering
         \includegraphics[width=0.99\textwidth]{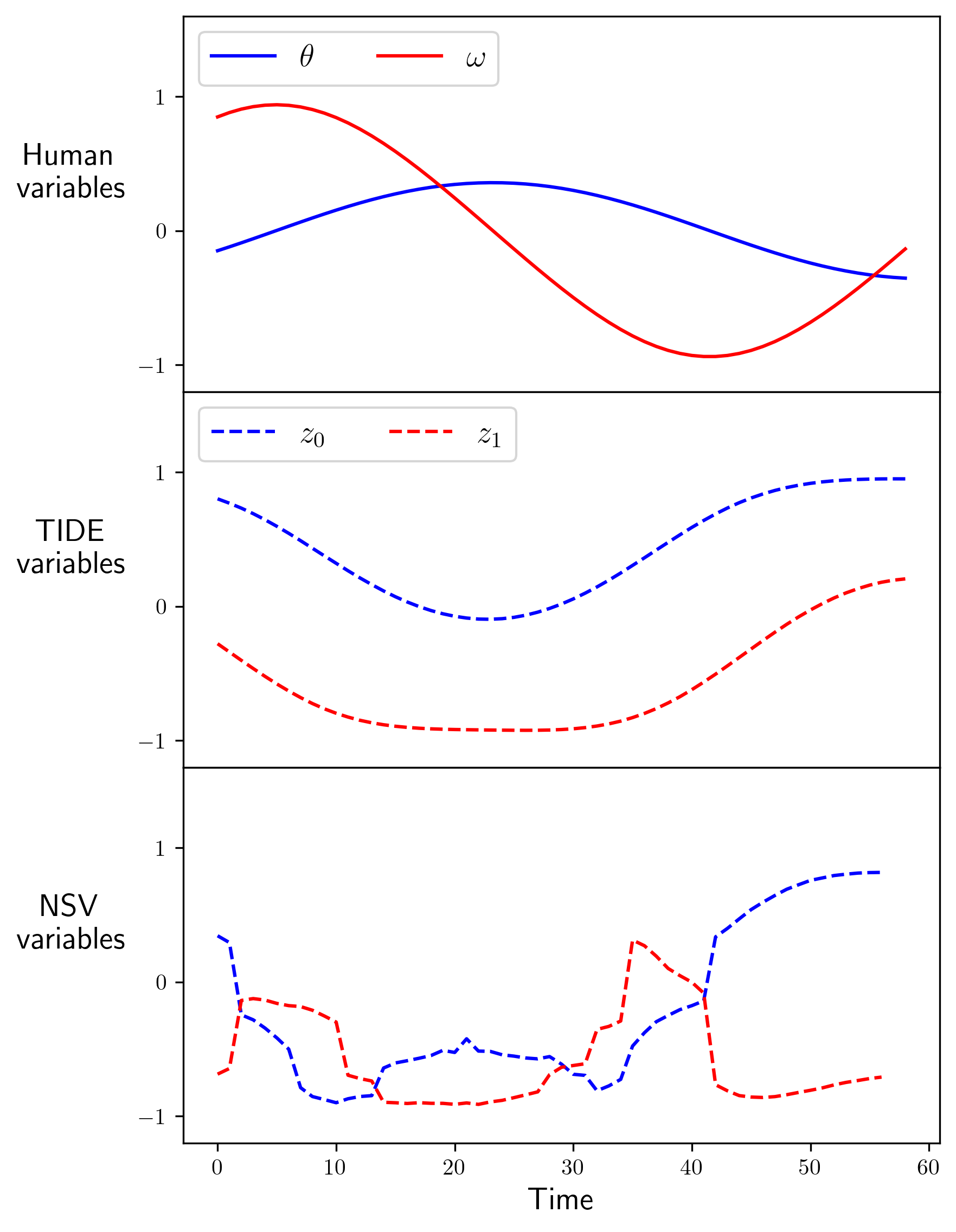}
         \caption{}
         \label{fig:single-pendulum-variables}
     \end{subfigure}
     \hfil
     \begin{subfigure}[t]{0.48\textwidth}
         \centering
         \includegraphics[width=0.99\textwidth]{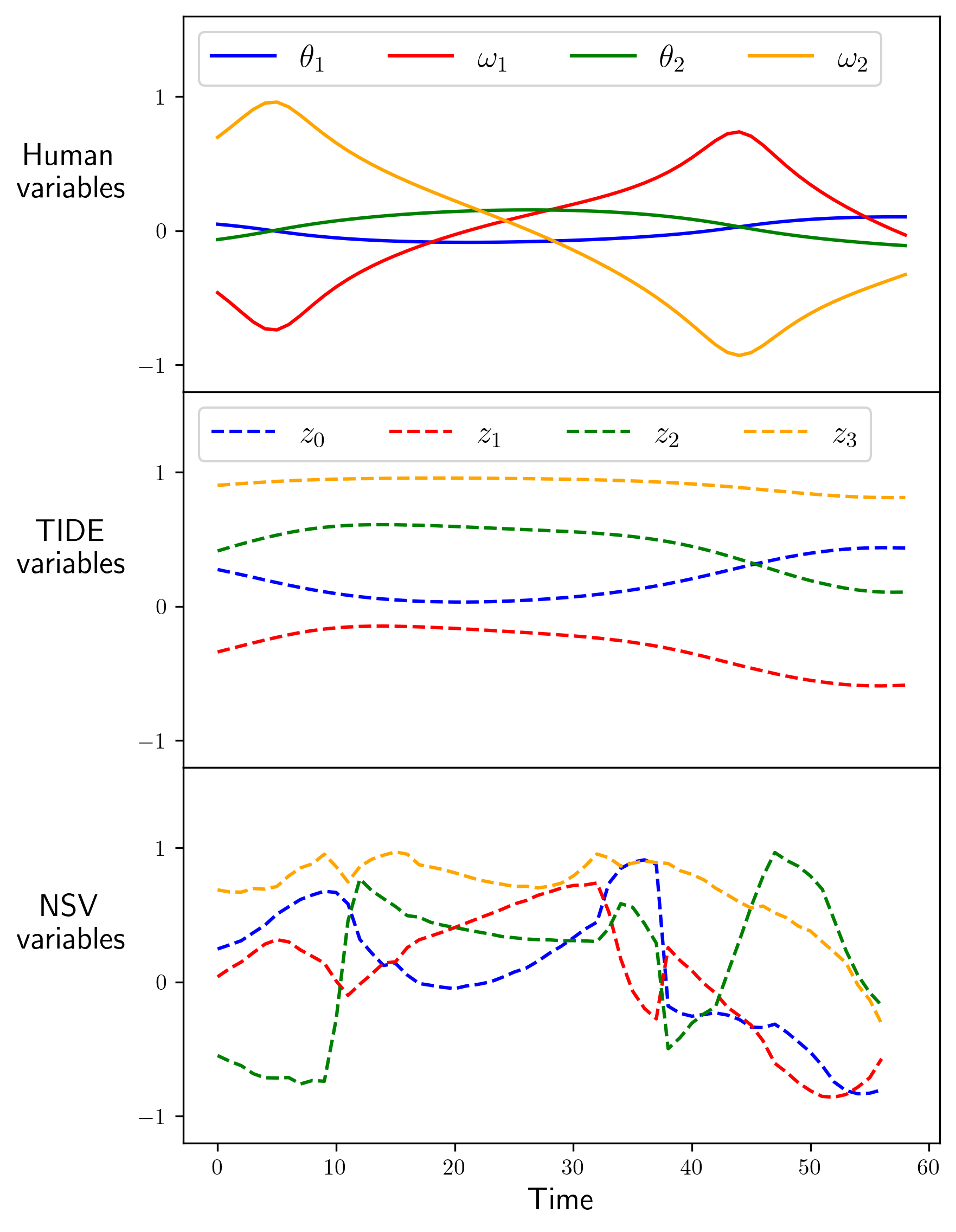}
         \caption{}
         \label{fig:double-pendulum-variables}
     \end{subfigure}
    \caption{Visual depiction of state variables for the single pendulum~\ref{fig:single-pendulum-variables} and double pendulum~\ref{fig:double-pendulum-variables} over a single video. Human variables are represented with solid lines, while model variables are shown with dashed lines. TIDE learns smoother variables that more closely resemble the angle and angular velocity compared to the baseline. We also depict the symbolic regression fit for the first latent variable in a solid black line. The analytical expressions are provided in Equations~\ref{eq:single-pendulum-formula} and~\ref{eq:double-pendulum-formula}.}
    \label{fig:state-variables}
\end{figure*}

\begin{figure}[t!]
    \centering
    \includegraphics[width=0.88\linewidth]{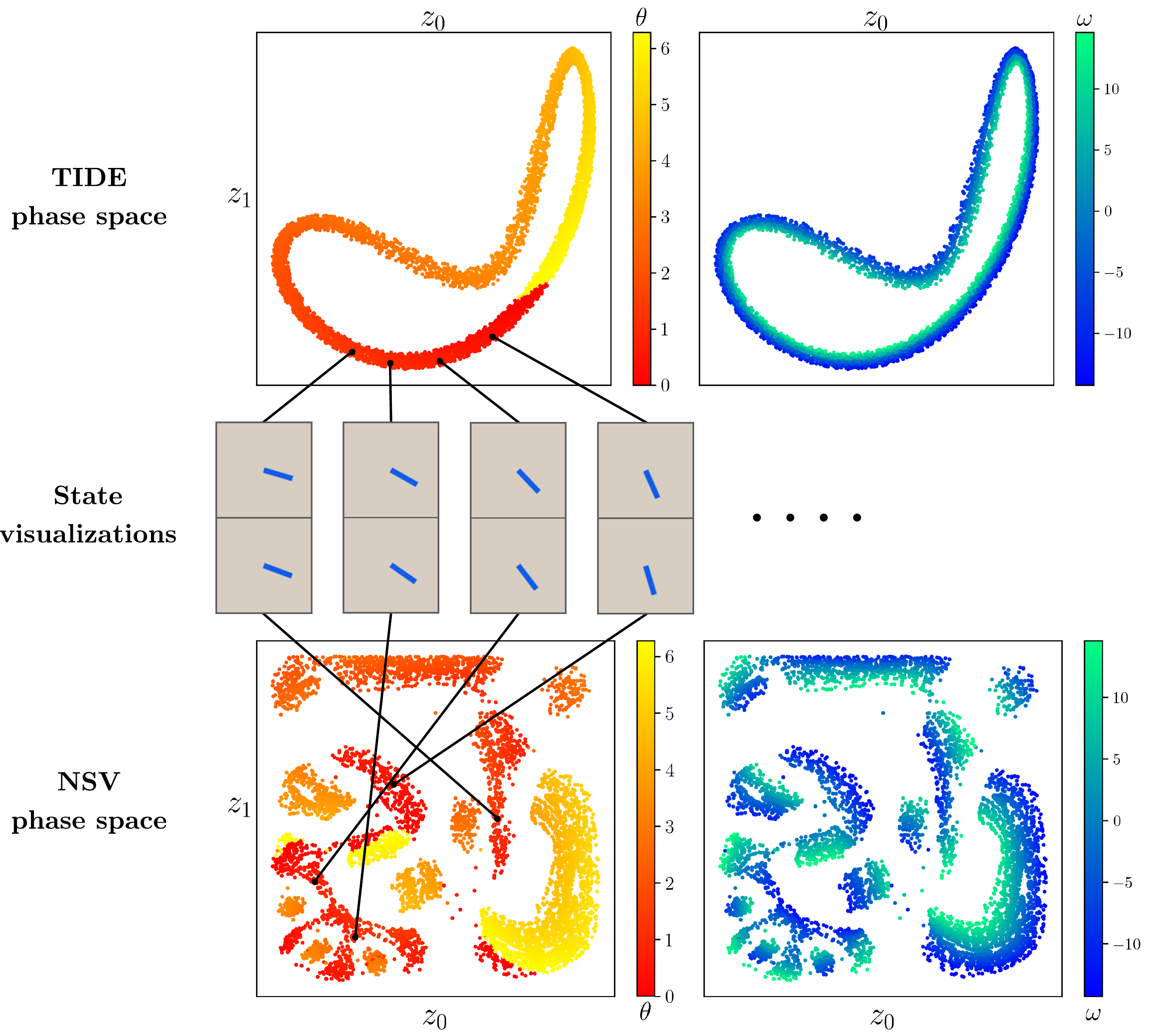}
    \caption{Phase space of the model variables on the single pendulum dataset, colored by the angle $\theta$ and angular velocity $\omega$. TIDE (top) generates a continuous latent space, a result of the time-derivative regularization. In contrast, the baseline (bottom) shows discontinuous transitions between consecutive states, which is less aligned with physical intuition.}
    \label{fig:phase-space}
\end{figure}

\subsection*{State variable visualization}
We present the learned latent space and state variables, and qualitatively compare them to the baseline approach. Figure~\ref{fig:state-variables} shows the state variables over a single video for both the single and double pendulum. Our learned variables closely resemble human-interpretable counterparts by maintaining continuity over time, while the baseline approach exhibits discontinuous jumps throughout the video. Such behavior diverges from typical human variables, like angle and angular velocity, which do not experience abrupt changes between consecutive observations. Additional visualizations of state variables for other datasets can be found in the Supplemental Information.

Our state variables not only exhibit qualitative similarity to their human equivalents but also capture meaningful information about observable physical phenomena. For example, in the case of the single pendulum, the angle of the pendulum arm is inversely correlated with the value of $z_0$: the angle reaches its maximum when $z_0$ is at its minimum, and vice versa. It is important to note that our representation does not directly correspond to standard human quantities, as state variables are not unique. Rather than explicitly predicting human variables, our goal is to align the human and AI representations as closely as possible, without relying on prior knowledge of the system.

For the circular motion and single pendulum datasets, we analyze the phase space of the state variables in Figure~\ref{fig:phase-space}. Each phase space is visualized using two colored point sets, where the colors represent the values of the human variables. Our model generates a continuous latent space, with angular and radial directions corresponding to the angle and angular velocity, respectively. In contrast, the baseline model produces a fragmented phase space. This difference is primarily due to the regularization penalty in our approach, which reduces discontinuities in the phase space by minimizing discrete derivatives. To demonstrate this effect, we show state variable transitions over four observations within a 10-frame segment. Our variables evolve smoothly over time, whereas the baseline model exhibits abrupt changes across different regions in the latent space.

\begin{table*}[t!]
    \centering
    \footnotesize
    \caption{Smoothness and correlation metrics between human and model variables across nine benchmark datasets. Higher values for MI, and lower values for smoothness and AMSE are preferred, as indicated by the arrows. Bold entries highlight the best-performing model for each metric and dataset, while missing values indicate cases where standard human variables are unavailable. TIDE learns smoother state variables that exhibit a stronger correlation with the angle and angular velocity compared to the baseline.}

    \renewcommand{\arraystretch}{1.5}
    \begin{tabular}{c@{\hskip 0.18in}cc@{\hskip 0.18in}cc@{\hskip 0.18in}cc}
    \hline \\ [-3.0ex]
    \multicolumn{1}{c}{\multirow{2}{*}{\textbf{Dataset}}} & \multicolumn{2}{c}{\textbf{Smoothness ($\downarrow$)}} & \multicolumn{2}{c}{\textbf{MI ($\uparrow$)}} & \multicolumn{2}{c}{\textbf{AMSE ($\downarrow$)}} \\ \cmidrule(l){2-7} \multicolumn{1}{c}{} & TIDE & NSV & TIDE & NSV & TIDE & NSV \\
    \hline \\ [-2.8ex]
    Reaction diffusion  & $\underline{\mathbf{1.16 \pm 0.06}}$ & $94.85 \pm 9.86$ & & \\
    Circular motion & $\underline{\mathbf{2.52 \pm 0.08}}$ & $438.09 \pm 41.76$ & $\underline{\mathbf{0.44 \pm 0.09}}$ & $0.30 \pm 0.01$ & $\underline{\mathbf{0.005 \pm 0.002}}$ & $0.34 \pm 0.018$ \\
    Single pendulum & $\underline{\mathbf{1.13 \pm 0.06}}$ & $177.78 \pm 29.32$ & $\underline{\mathbf{0.68 \pm 0.04}}$ & $0.48 \pm 0.01$ & $\underline{\mathbf{0.006 \pm 0.001}}$ & $0.34 \pm 0.033$ \\
    Double pendulum & $\underline{\mathbf{1.84 \pm 0.18}}$ & $52.03 \pm 12.93$ & $\underline{\mathbf{1.39 \pm 0.01}}$ & $1.35 \pm 0.10$ & $\underline{\mathbf{0.007 \pm 0.002}}$ & $0.26 \pm 0.019$ \\
    Elastic pendulum & $\underline{\mathbf{4.89 \pm 0.46}}$ & $35.62 \pm 0.77$ & $\underline{\mathbf{2.08 \pm 0.18}}$ & $2.05 \pm 0.08$ & $\underline{\mathbf{ 0.012 \pm 0.007}}$ & $0.20 \pm 0.002$ \\
    Swing stick & $\underline{\mathbf{6.26 \pm 0.75}}$ & $29.7 \pm 12.89$ & & \\
    Air dancer & $\underline{\mathbf{3.79 \pm 0.23}}$ & $83.27 \pm 41.0$ & & \\
    Lava lamp &  $\underline{\mathbf{26.95 \pm 1.17}}$ & $158.63 \pm 8.17$ & & \\
    Fire flame & $\underline{\mathbf{54.74 \pm 4.27}}$ & $415.45 \pm 35.45$ & & \\[0.5ex] \hline
    \end{tabular}
    \label{tab:metrics}
\end{table*}

\subsection*{Interpretability metrics}
TIDE learns a continuous latent space with smoother state variables that more closely resemble human quantities. We conduct further quantitative analysis to demonstrate a stronger correlation between human and AI variables, based on smoothness, mutual information (MI), and analytical mean squared error (AMSE).

Table~\ref{tab:metrics} reports smoothness penalty ($\mathcal{L}_\text{reg}$), MI, and AMSE values for the circular motion, single pendulum, and double pendulum datasets. Across all systems, our variables are consistently smoother compared to the baseline. Additionally, the MI between TIDE and human variables is higher, indicating a stronger correlation. By employing symbolic regression to derive an analytical expression for the latent representation, we achieve significantly lower AMSE, further validating the use of smoothness as a proxy for interpretability when prior knowledge of the system or its dynamics is unavailable.

In Figure~\ref{fig:state-variables}, we present the state variables and their analytical fits over a single video for the single and double pendulum datasets. We accurately derive an expression for the TIDE variables using symbolic regression, where the first latent component for the single and double pendulum is respectively described as:


\begin{align}
    z_0 ={}& (0.27 - 0.15 \sin(\theta + 1.48)) \sin(2\theta - 0.9) \notag \\
         & {}- 0.74\sin(\theta) - 0.26,
         \label{eq:single-pendulum-formula} \\
    z_0 ={}& (0.2\sin(\theta_1 - 0.34) + 0.47) \sin(\theta_1 - 0.17) \notag \\
         & {}+ 0.047(\sin(\theta_1 + 0.6) + 1) \sin(\theta_2 - 0.28) \notag \\
         & {}- 0.67.
         \label{eq:double-pendulum-formula}
\end{align}

In contrast, no symbolic combination of elementary functions, angle, and angular velocity succeeds in capturing information about the baseline variables. Additional analytical fits and equations are included in the Supplemental Information.
\section{Conclusion}\label{sec:conclusion}
Automated scientific discovery is becoming increasingly feasible with the rise of data-driven methodologies. This process involves extracting descriptive variables from observations and formulating new laws and equations based on these variables. Traditionally, identifying state variables has relied on human expertise, but recent advancements aim to automatically discover variables from raw data without prior knowledge of the system. However, previous approaches have struggled to produce human-interpretable physics due to the non-uniqueness of state representations. While identifying state variables is crucial, it is equally important to prioritize those that offer interpretive value for scientific discovery.

In this work, we introduce a novel Temporally-Informed Dynamics Encoder (TIDE) designed to derive meaningful state variables from observational data without requiring prior knowledge of the system, statistical relationships between variables, or parametric assumptions for the dynamics. Our approach emphasizes aligning AI representations with human-interpretable state variables based on three key desired characteristics. These properties are incorporated into an autoencoder framework by jointly optimizing a dynamics module and applying time-derivative regularization. TIDE significantly outperforms the existing neural state variable baseline by learning variables with meaningful physical significance and stronger correlation to human quantities. Furthermore, we derive analytical expressions for the predicted variables using symbolic regression. To the best of our knowledge, this is the first result to derive explicit formulas for state variables learned \textit{without prior knowledge of the system}.

A current limitation of our method is its reliance on the stable state variable representation assumption, which may not apply in scenarios where data collection is inefficient. One potential direction for enhancing the interpretive value of learned variables in broader settings is to incorporate semi-supervised or curriculum learning. In this approach, knowledge of intuitive human variables, such as angle or speed, could be transferred to novel environments where similar underlying variables may be hidden in the data. Although our method aligns model and human variables by encouraging specific properties in the latent space, the resulting representation may not perfectly match human quantities, given the non-uniqueness of state variable configurations. Another promising avenue is to theoretically explore sufficient conditions on the data-generating process under which state variables become identifiable. 
\section{Acknowledgements}\label{sec:acknowledgements}
This research was supported in part by NSF AI Institute for Dynamical Systems \#2112085, DARPA MTO Lifelong Learning Machines (L2M) Program HR0011-18-2-0020,
NSF NRI \#1925157, NSF DMS \#1937254, NSF DMS \#2012562, NSF CCF \#1704833, and DE \#SC0022317. The code and implementation for our work is available at \url{https://github.com/kzhangm02/tide}.

\bibliography{example_paper}
\bibliographystyle{icml2026}

\newpage
\appendix
\onecolumn
\appendix
\newpage

\section{Qualitative reconstructions} \label{app:reconstructions}
\begin{figure}[h]
    \centering
     \begin{subfigure}[t]{0.45\textwidth}
         \centering
         \includegraphics[width=\textwidth]{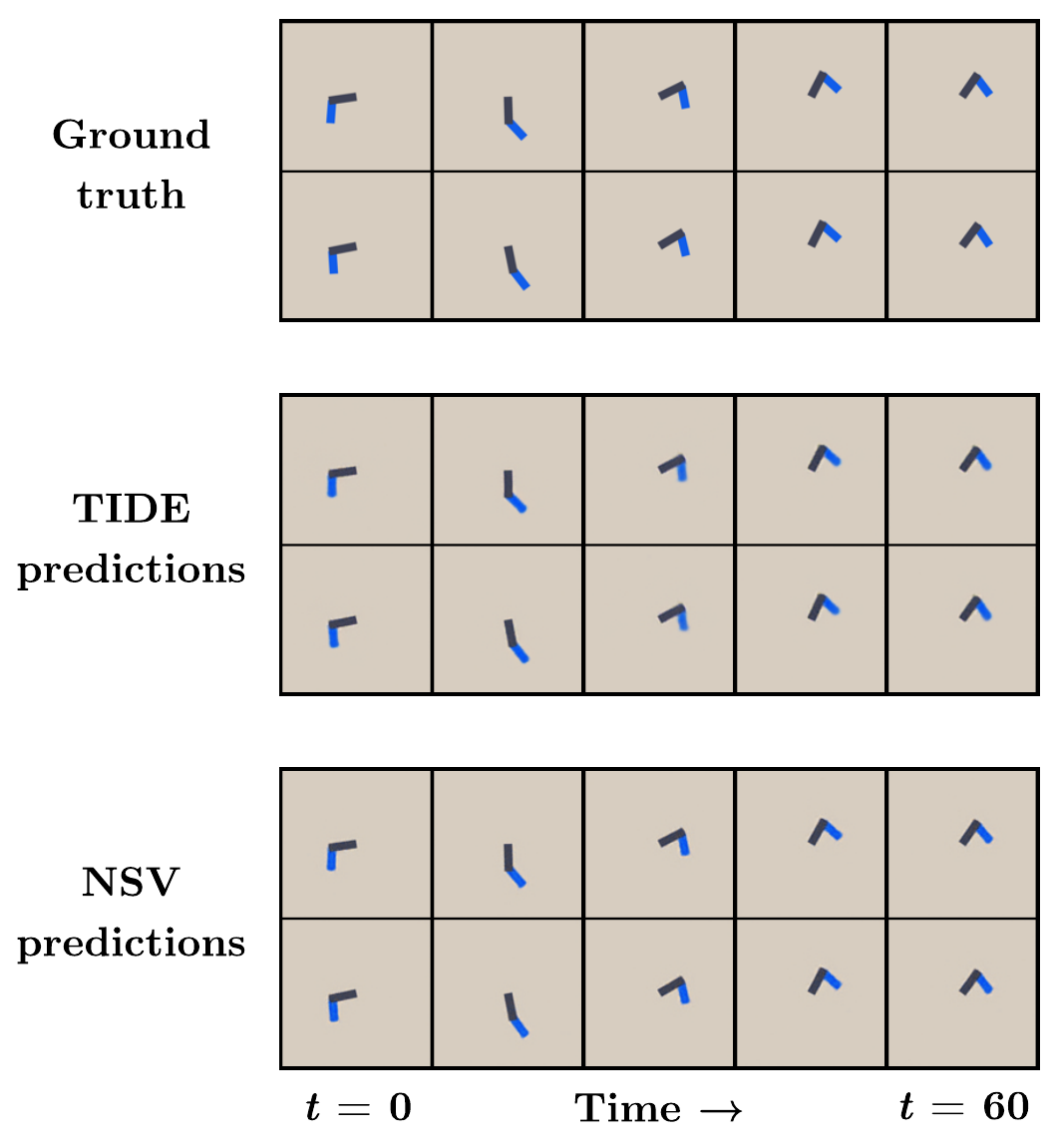}
         \caption{}
         \label{fig:reconstruction-double-pendulum}
     \end{subfigure}
     \hfil
     \begin{subfigure}[t]{0.45\textwidth}
         \centering
         \includegraphics[width=\textwidth]{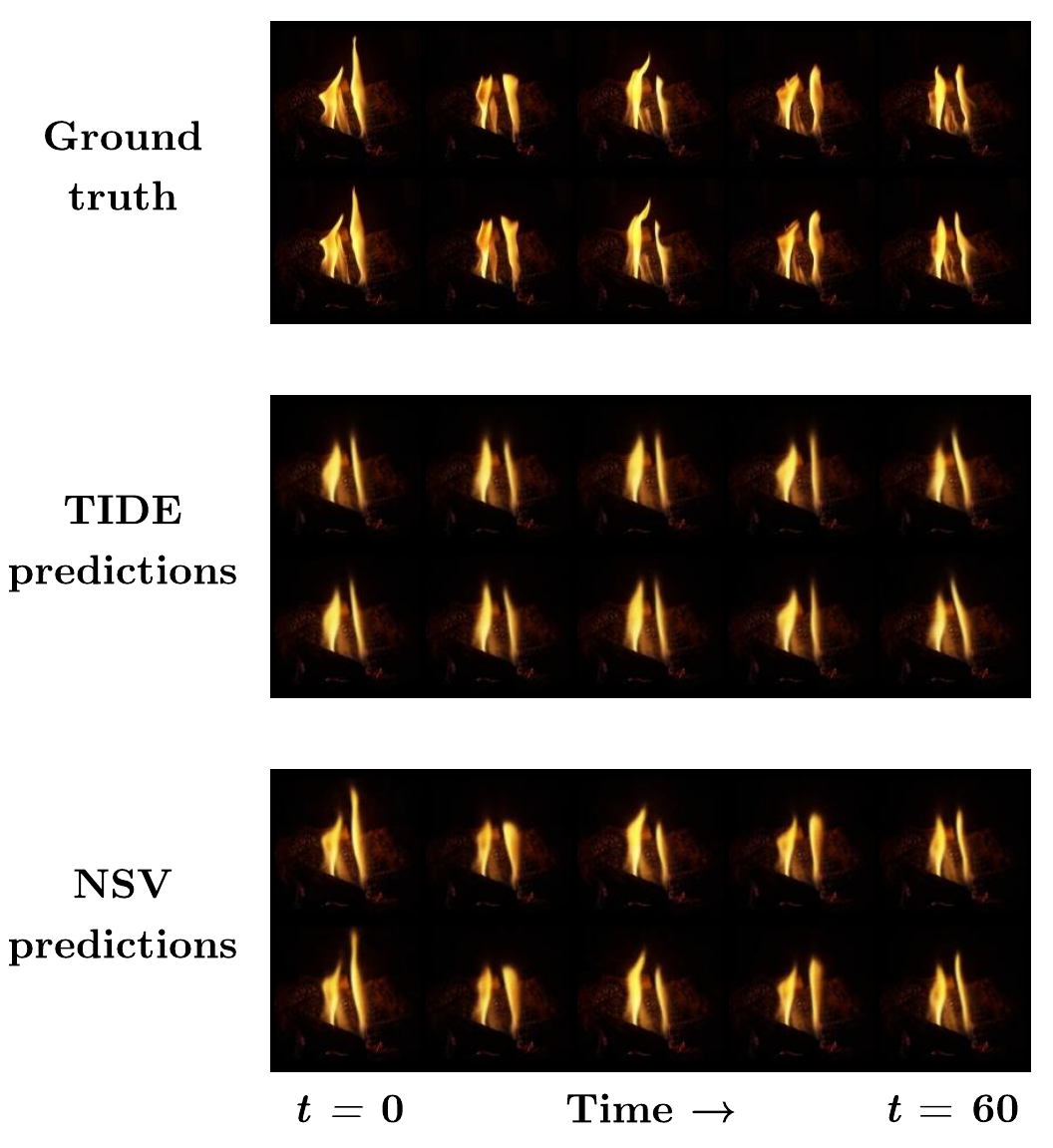}
         \caption{}
         \label{fig:reconstruction-fire-flame}
     \end{subfigure}
    \caption{Comparison of reconstructed states for the double pendulum (a) and fire flame (b) datasets. TIDE successfully captures the key physical elements and dynamics of the systems.}
    \label{fig:reconstructions}
\end{figure}

To assess reconstruction quality using TIDE, we compare the predicted images from our model and a baseline approach in Figure~\ref{fig:reconstructions}. Specifically, we include visualizations from the double pendulum and fire flame datasets, as these exhibit the largest reconstruction discrepancy. In the double pendulum case, the additional error largely results from distortions in the pendulum arm, although key physical attributes such as angle and angular velocity are still accurately captured. Similarly, for the fire flame dataset, our model generates images with some added blur but successfully reconstructs the shape and orientation of the flames. These artifacts may be reduced through improved hyperparameter selection. Overall, TIDE networks are capable of learning system dynamics while simultaneously producing a more interpretable state variable representation.

\section{Additional state variable visualizations} \label{app:visualizations}
\begin{figure}[h]
    \centering
     \begin{subfigure}[t]{0.48\textwidth}
         \centering
         \includegraphics[width=\textwidth]{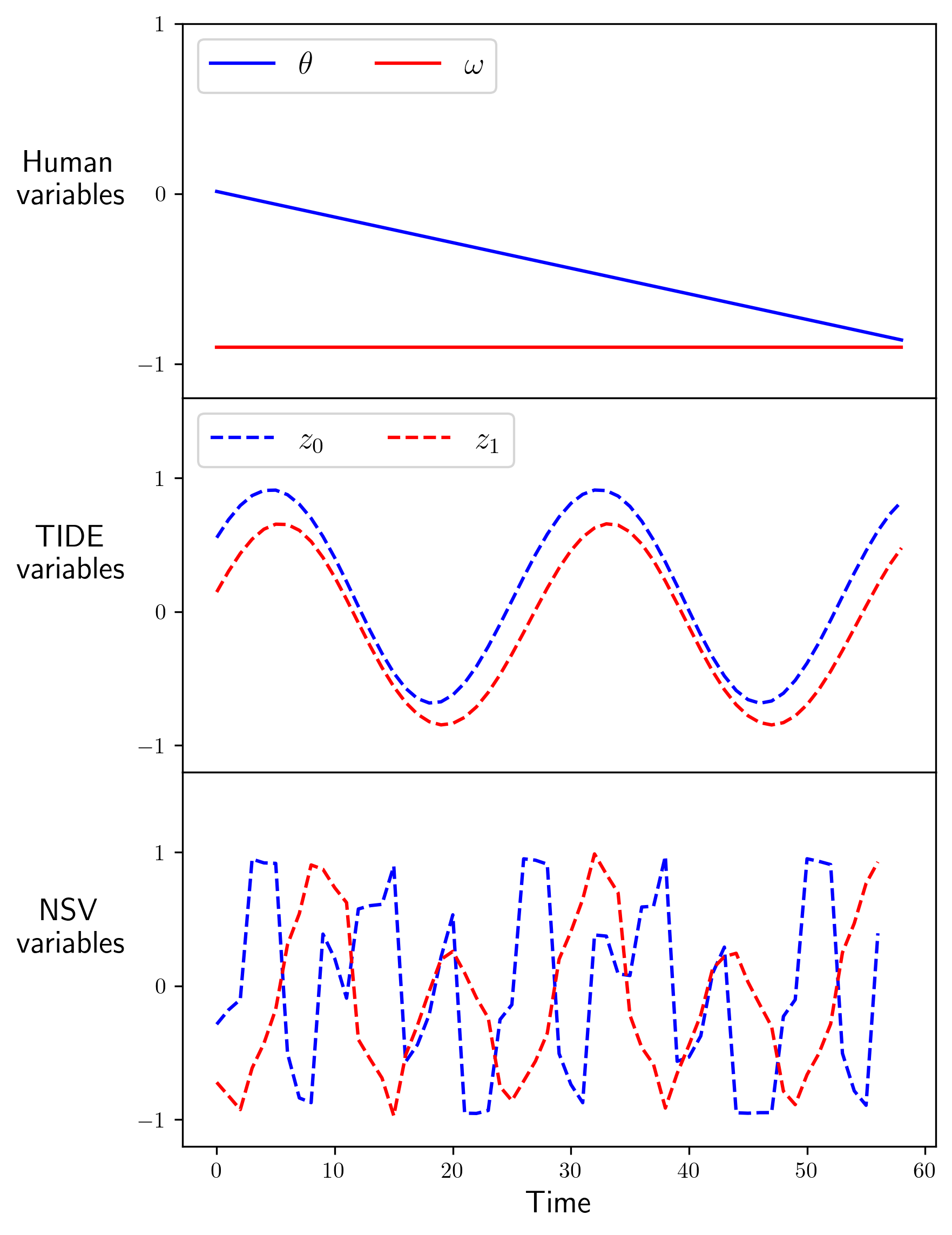}
         \caption{}
         \label{fig:circular-motion-variables}
     \end{subfigure}
     \hfil
     \begin{subfigure}[t]{0.48\textwidth}
         \centering
         \includegraphics[width=\textwidth]{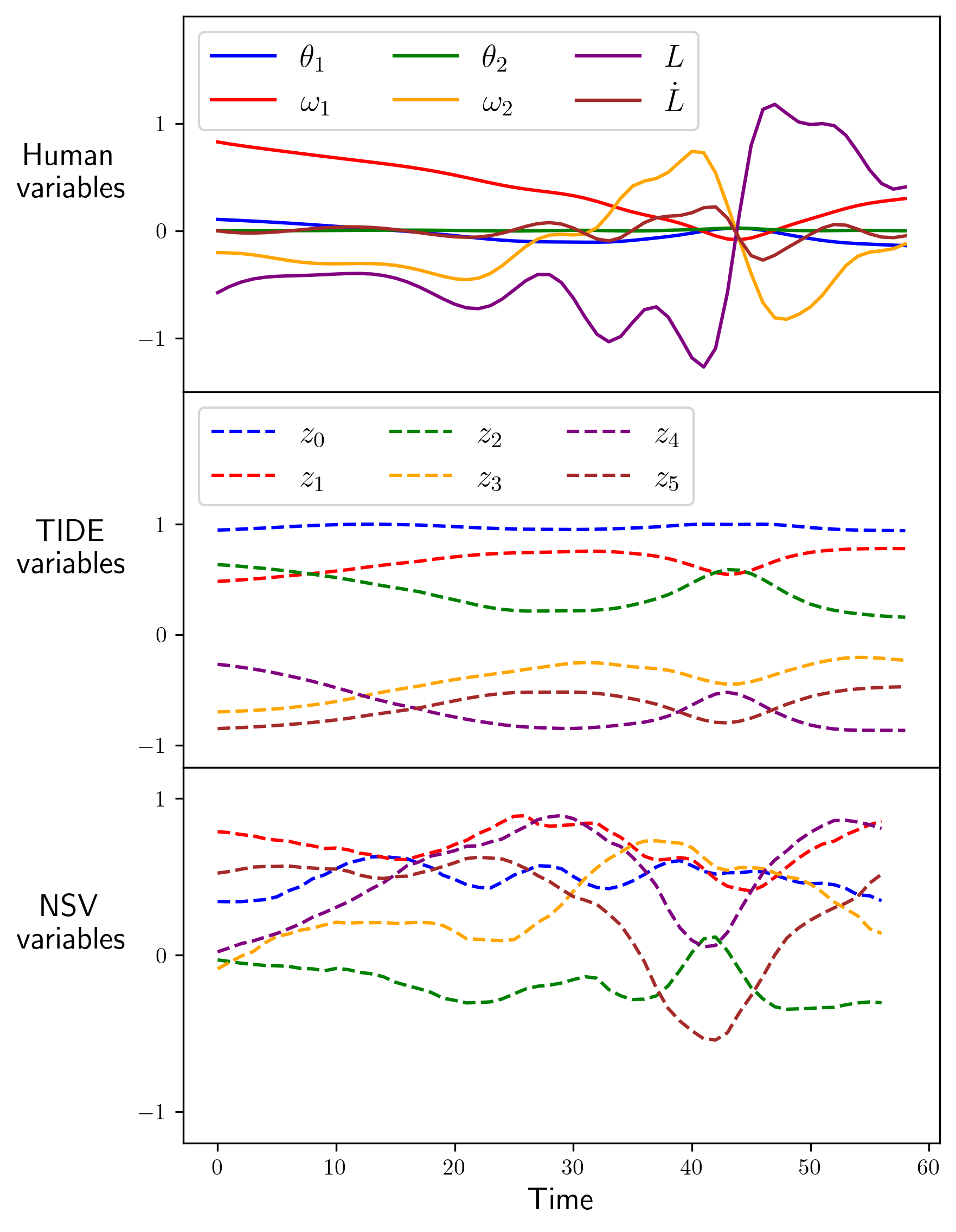}
         \caption{}
         \label{fig:elastic-pendulum-variables}
     \end{subfigure}
    \caption{Visual depiction of state variables for the circular motion (a) and elastic pendulum (b) across a single video. Standard human variables are represented by solid lines, while the model-predicted variables are shown with dashed lines. In the case of the elastic pendulum, $L$ denotes the length.}
    \label{fig:state-variables-app}
\end{figure}

\begin{figure}[h]
    \centering
    \includegraphics[width=0.86\textwidth]{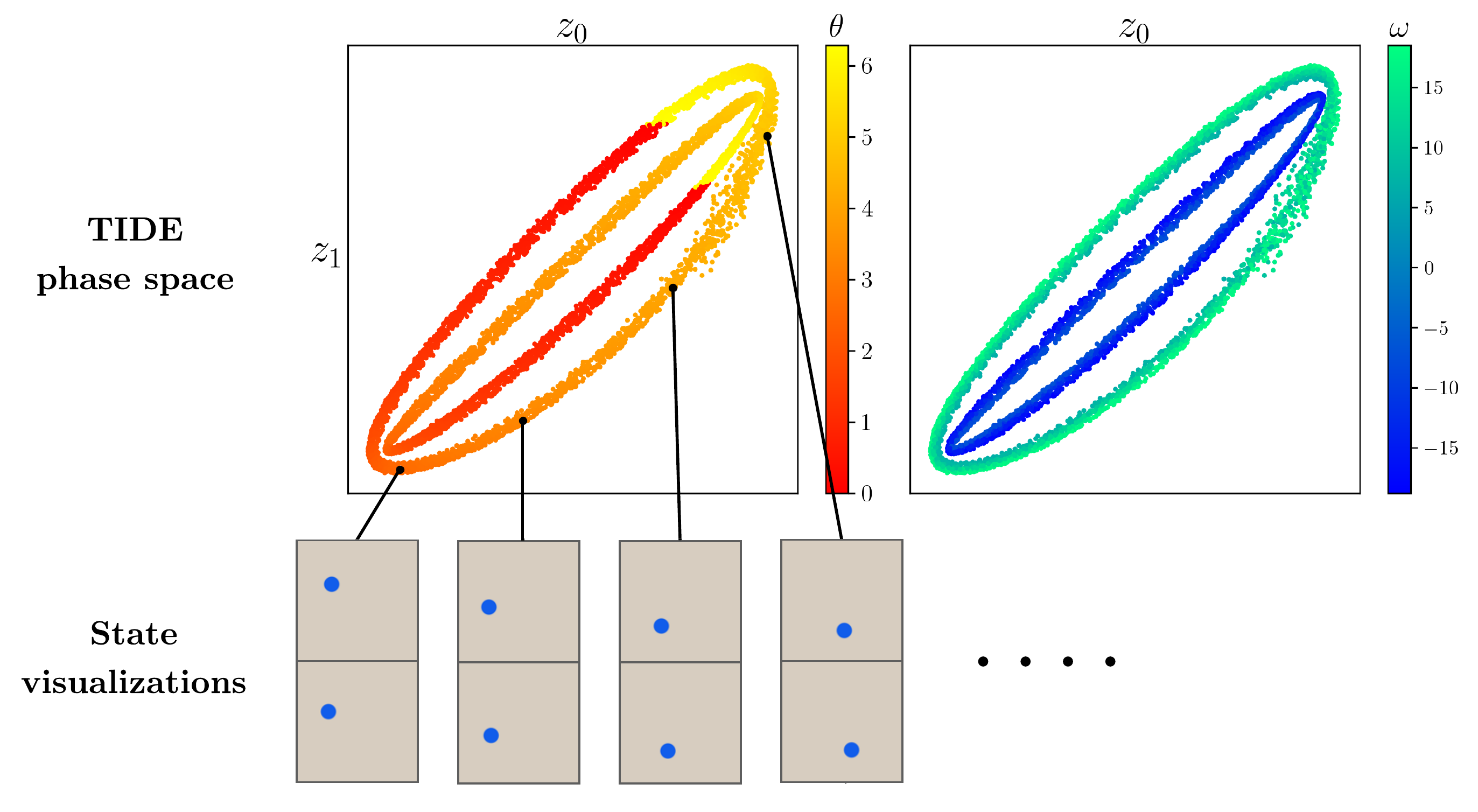}
    \caption{Phase space of TIDE state variables for the circular motion dataset, colored by angle $\theta$ and angular velocity $\omega$. Four states from a 10-frame segment are visualized to illustrate the effect of regularization.}
    \label{fig:phase-space-app}
    \vspace{8pt}
\end{figure}

We illustrate the state variables for the circular motion and elastic pendulum datasets in Figure~\ref{fig:state-variables-app}. Similar to the single and double pendulum scenarios, TIDE learns variables that are smoother and more closely aligned with human-interpretable quantities. Additionally, Figure~\ref{fig:phase-space-app} shows the phase space of the circular motion dataset, consisting of two distinct ellipses that correspond to positive and negative angular velocity. This observation further supports TIDE's ability to capture physically meaningful quantities. To demonstrate the effect of time-derivative regularization, we include visualizations of four states across a 10-frame segment, yielding results consistent with those observed for the single pendulum. Since the angular velocity remains constant throughout each video, the state variables evolve smoothly over time.

\section{Additional state variable analytical fits}
\label{app:analytical_fit}
\begin{figure}[h!]
    \centering
    \begin{subfigure}[t]{0.47\textwidth}
        \centering
        \includegraphics[width=\textwidth]{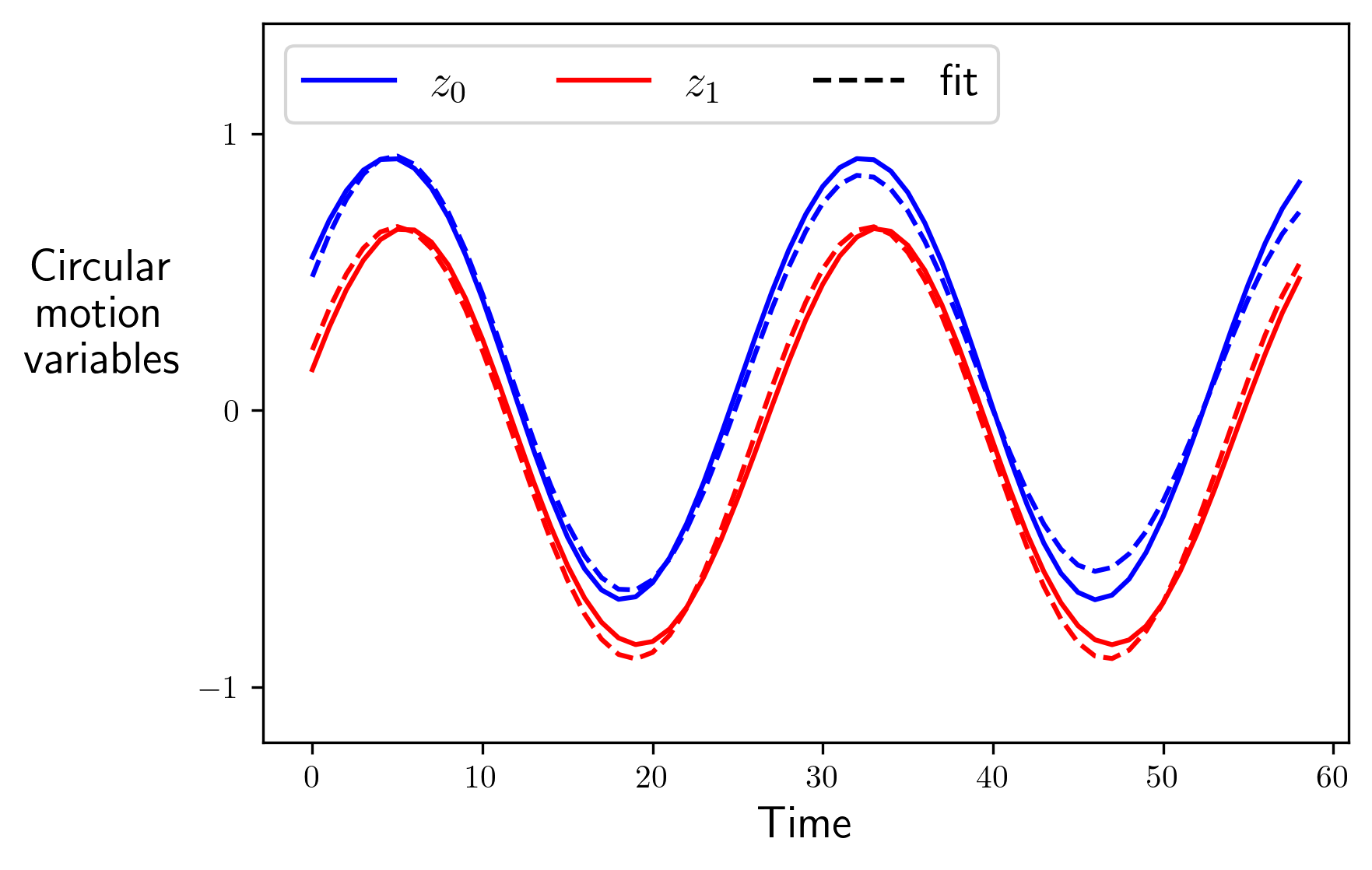}
        \caption{}
        \label{fig:circular-motion-full-fit}
    \end{subfigure}
    \hfil
    \begin{subfigure}[t]{0.47\textwidth}
        \centering
        \includegraphics[width=\textwidth]{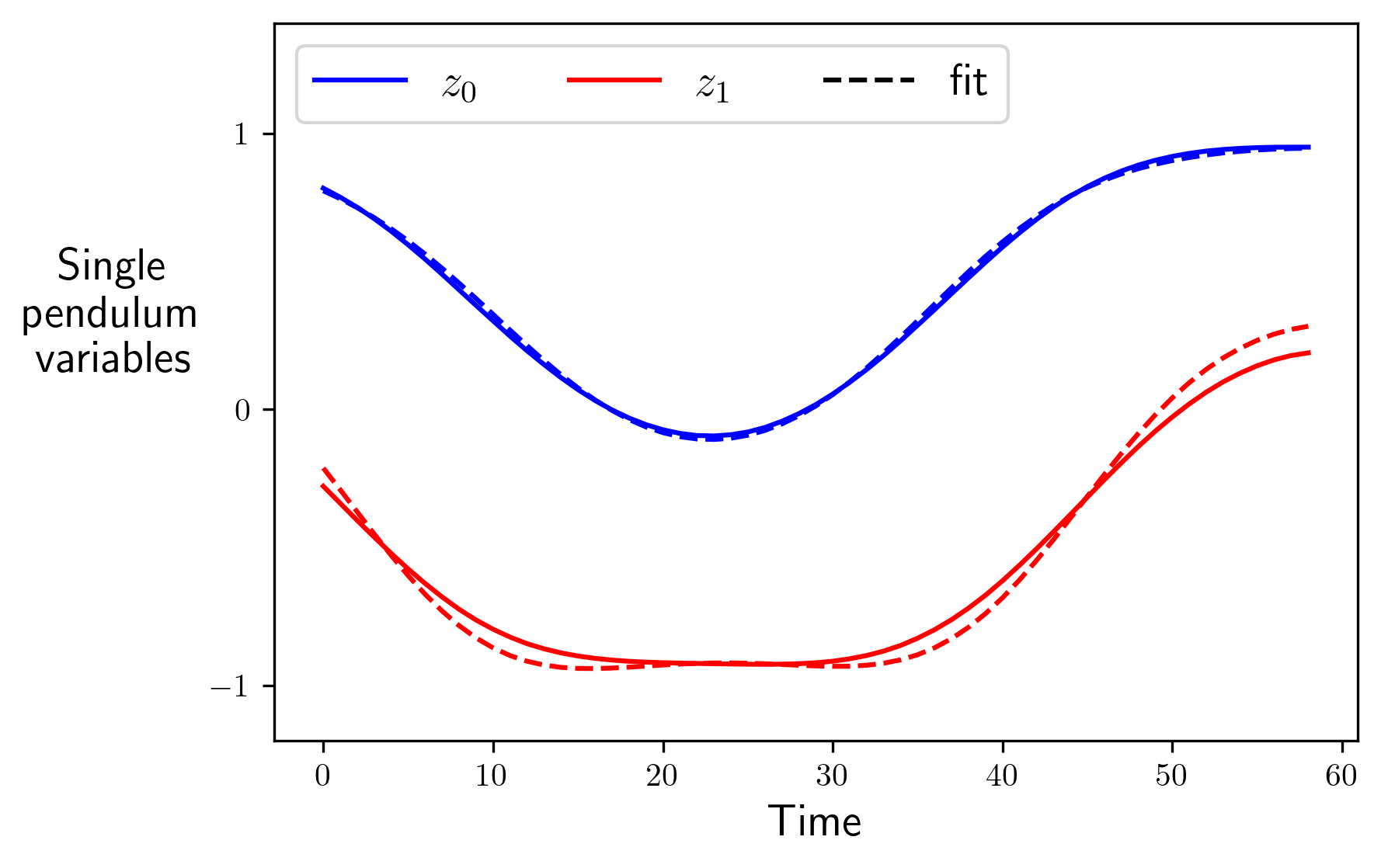}
        \caption{}
        \label{fig:single-pendulum-full-fit}
    \end{subfigure}
    \hfil
    \begin{subfigure}[t]{0.47\textwidth}
        \centering
        \includegraphics[width=\textwidth]{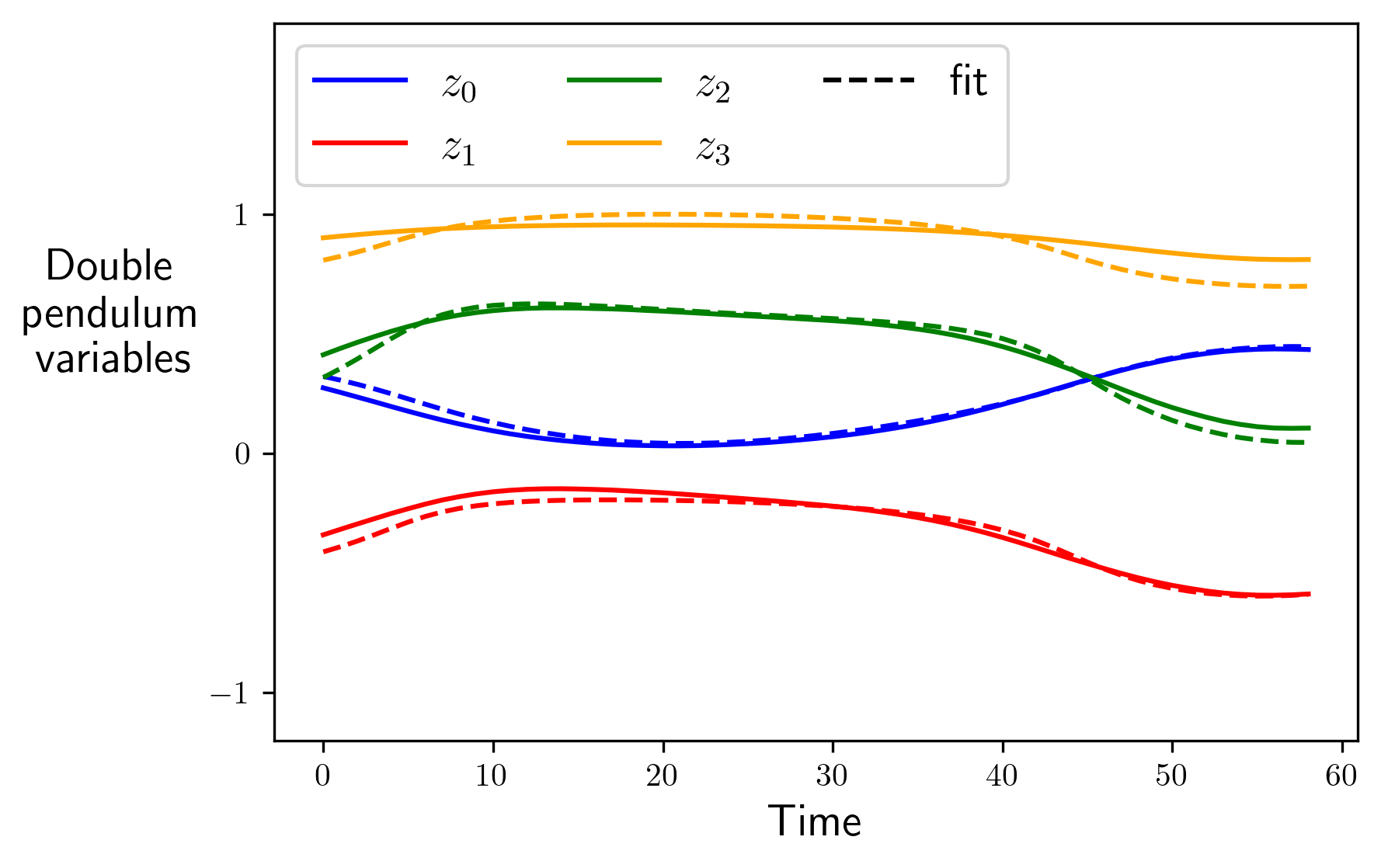}
        \caption{}
        \label{fig:double-pendulum-full-fit}
    \end{subfigure}
    \hfil
     \begin{subfigure}[t]{0.47\textwidth}
         \centering
         \includegraphics[width=\textwidth]{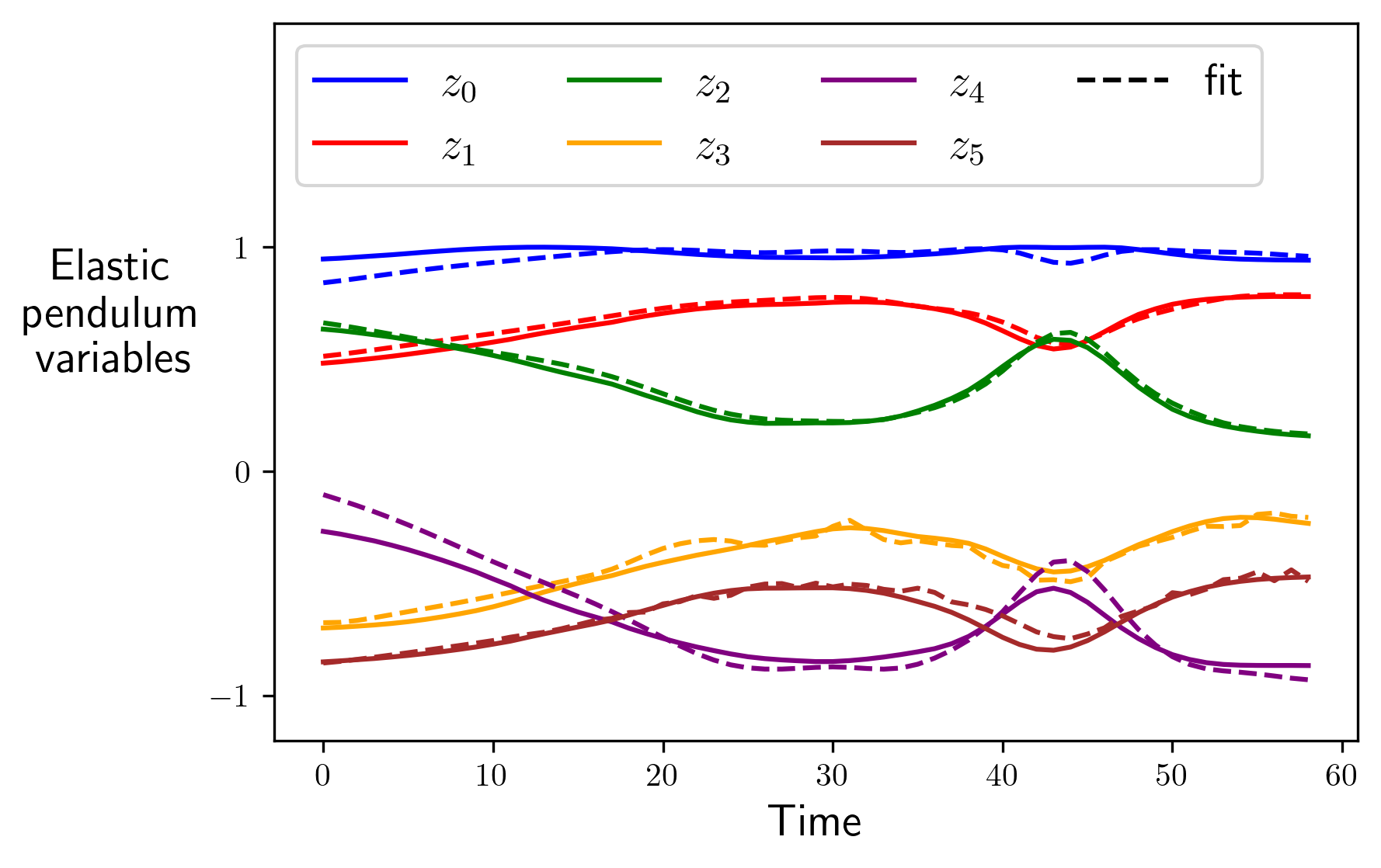}
         \caption{}
         \label{fig:elastic-pendulum-full-fit}
     \end{subfigure}
     \hfil
    \caption{Analytical fits for all TIDE state variables for the circular motion (a), single pendulum (b), double pendulum (c), and elastic pendulum (d) datasets. The model-predicted variables are shown as solid lines, while their corresponding analytical fits are represented by dashed lines.}
    \label{fig:full-analytical-fits}
\end{figure}

We present the analytical fit of all TIDE state variables for the circular motion, single pendulum, double pendulum, and elastic pendulum datasets in Figure~\ref{fig:full-analytical-fits}. Our model accurately predicts state variables expressed in terms of angle and angular velocity, as the learned representation aligns closely with human-interpretable quantities. For conciseness, we omit the analytical fit from the baseline model, as its behavior mirrors the NSV results shown in Figure 2.

\section{State variable equations} 
\label{app:equations}

We present the analytical expressions for TIDE variables obtained through symbolic regression in Table~\ref{tab:equations}. To the best of our knowledge, this is the first instance of deriving explicit formulas for state variables learned \textit{without prior knowledge of the system}.
\begin{table}[h]
    \renewcommand{\arraystretch}{1.5}
    \centering
    \begin{tabular}{c||l}
        Dataset & \multicolumn{1}{c}{Equations}  \\
        \hline\hline
        \multirow{2}{*}[-0.4em]{\shortstack{Circular\\motion}} & $z_1 = -0.64 \sin(\theta) + 0.5 \cos(1.03 \theta) - 0.5 \cos(0.018 \omega) + 0.6$ \\
        & $z_2 = 0.0068 \omega \sin(\theta + 0.96) - 0.79 \sin(\theta - 0.75) - 0.12$ \\ & \\[-3.0ex]
        \hline 
        
        \multirow{2}{*}[-0.4em]{\shortstack{Single\\pendulum}}  & $z_1 = 0.26 (0.0085 \omega - 0.52 \sin(\theta) - \cos(\theta) - 0.58)^2 - \sin(\theta)$ \\
        & $z_2 = - 0.6 \sin(\theta + 0.27) - (\sin(\theta) + \sin(0.0012 \omega) - \sin(\theta - 0.72))^2$ \\ & \\[-3.0ex]
        \hline & \\[-3.0ex]
        
        \multirow{4}{*}[-0.4em]{\shortstack{Double\\pendulum}} & $\begin{aligned} z_1 ={}& 0.38 \sin(\theta_1 - 0.85)^2 + 0.059 \sin(\theta_1 + 0.66)^2 + 0.55 \sin(\theta_1) \end{aligned}$ \\
        & $z_2 = -0.12 \sin(\theta_2) - \sin(0.52 \theta_1 + 0.51)^2 - 0.057$ \\
        & $z_3 = -0.00021 \sin(\omega_2 (\omega_1 - \theta_1 + 1.17))^2 - 0.66 \sin(\theta_1 - 0.58) - 0.18 \sin(\theta_2 + 5.3)$ \\
        & $z_4 = 0.00063 (\omega_2 - \omega_1) - 0.0013 \theta_2 - 0.25 \sin(\theta_1 - 0.48) + 0.77$ \\ & \\[-3.0ex]
        \hline & \\[-3.0ex]
        
        \multirow{6}{*}[-1.2em]{\shortstack{Elastic\\pendulum}} & $z_1 = -(\theta_2 + 0.21) \sin(\theta_1)^2 - (\sin(\theta_1) - 0.0076) (\sin(\theta_2) + 0.12) + 0.97$ \\
        & $\begin{aligned} z_2 ={} & -(\sin(1.05 \theta_1) \sin(\theta_2 + 0.29) + 0.49)^2 + 0.073 \sin(0.075 \omega_1^2) \\
        & - 0.011 \sin(\omega_2)^2 \sin(L)^2 \sin(\theta_2^2 \dot{L}^2) + 0.87 \end{aligned}$ \\
        & $\begin{aligned} z_3 ={} & 0.93 (\sin(\theta_2) + 0.45) \sin(\theta_1 + \theta_2 + 0.062) + 0.41 \end{aligned}$ \\
        & $\begin{aligned} z_4 ={} & - 0.0036 \omega_2 + (\theta_2 + 0.23) (\theta_2 - 1.8 \sin(\theta_1)) \\
        & - 0.0037 (\sin(\omega_1) + \sin(L) + 3.03)^2 - 0.43 \end{aligned}$ \\
        & $\begin{aligned} z_5 ={} &  0.47 \theta_2 + 0.0012 \omega_2 - 0.00035 \dot{L} \sin(\omega_1 + L^2) - 0.0012 (\dot{L} - 0.48)^2 \\
        & + 0.47 (\sin(\theta_1 + 0.25) + 0.64)^2 - 0.94 \end{aligned}$ \\
        & $z_6 =  -(\sin(\theta_2) + 0.32) \sin(\theta_1) + 0.07 \sin(\omega_1 \omega_2 (L - \dot{L}))^2 \sin(\theta_1 - 0.4)^2 - 0.69$ \\
    \end{tabular}
    \caption{Analytical expressions for all TIDE state variables in terms of standard human-interpretable variables for the circular motion, single pendulum, double pendulum, and elastic pendulum datasets. For compactness, all constants and coefficients smaller than $10^{-4}$ are omitted.}
    \label{tab:equations}
\end{table}

\end{document}